\pdfoutput=1

\documentclass[11pt,a4paper]{article}
\usepackage[hyperref]{rocling2023}

\roclingfinalcopy 
\def\roclingpaperid{7} 

\usepackage{times}
\usepackage{latexsym}

\usepackage[T1]{fontenc}

\usepackage[utf8]{inputenc}

\usepackage{microtype}

\usepackage{booktabs}
\usepackage{multirow}

\usepackage{CJKutf8}

\usepackage{amsmath}
\usepackage{graphics}
\usepackage{graphicx}
\graphicspath{ {./figures/} }
\usepackage{subcaption,ragged2e}
\usepackage{comment}
\usepackage{color}
\usepackage{dsfont}

\usepackage{enumitem}
\usepackage{soul}
\newcommand{\Note}[1]{}
\renewcommand{\Note}[1]{\hl{[#1]}}  

\newcommand\B{\rule[-1.5ex]{0pt}{0pt}}
\usepackage{amsfonts,amssymb,bm,bbm,mathtools,upgreek}
\usepackage[all]{nowidow}
\def\nCls{K}

\def\dataSize{N}
\def\binSize{M}
\def\inpSet{\mathcal{X}}

\def\labSet{\mathcal{Y}}
\def\dataSet{\mathcal{D}}
\def\langSet{\mathcal{T}}
\def\langSetEq{\langSet = \{\textrm{es, fr, id, ja, zh}\}}
\def\tranDataSet{\widetilde{\dataSet}}
\def\tranDataSetEq{\tranDataSet = \bigcup_{t\in\langSet}\tranDataSet_t}
\def\binSet{\mathcal{B}}
\def\trainSet{\medmuskip=0mu \dataSet={\{(\inp_i,\lab_i)\}_{i=1}^{\dataSize}}}
\def\Real{\mathbb{R}}
\def\param{\theta}
\def\params{\bm{\param}}
\def\inp{x}
\def\inpTran{\tilde{\inp}}
\def\lab{y}
\def\predLab{\hat{\lab}}

\def\trueDist{q}
\def\predDist{p}
\def\predDistTran{\tilde{\predDist}}
\def\predDistEq{\predDist(\lab|\inp) = \softmax(\modelP(\inp))}

\def\conf{\hat{\predDist}}
\def\param{\theta}
\def\params{\bm{\param}}
\def\model{f}
\def\modelP{\model_{\params}}
\def\modelPeq{\modelP(\inp) = \MLP(\PLM(\inp))}

\def\inpVec{\mathbf{h}}
\def\inpTranVec{\tilde{\inpVec}}

\def\binId{i}
\def\sampId{j}

\def\binTh{\binId^{\textrm{th}}}

\def\binInterval{(\frac{\binId-1}{\binSize}, \frac{\binId}{\binSize}]}
\def\indicator{\mathbbm{1}}
\def\binAcc{\textrm{acc}(\binSet_\binId)}
\def\binConf{\textrm{conf}(\binSet_\binId)}
\def\ECEeqs{
\ECE     &= \sum^\binSize_{\binId=1}\frac{|\binSet_\binId|}{\dataSize}|\binAcc - \binConf|,\nonumber\\
\binAcc  &= \frac{1}{|\binSet_\binId|}\sum_{\sampId\in\binSet_\binId}\indicator(\predLab_\sampId = \lab_\sampId),\nonumber\\
\predLab_\sampId &= \argmax_{\lab_\sampId\in\labSet}\,\predDist(\lab_\sampId|\inp_\sampId),\nonumber\\
\binConf &= \frac{1}{|\binSet_\binId|}\sum_{\sampId\in\binSet_\binId}\conf_\sampId,\nonumber\\
\conf_\sampId &= \max_{\lab_\sampId\in\labSet}\,\predDist(\lab_\sampId|\inp_\sampId),\nonumber
}
\def\Pi{p_i}

\def\max{\textrm{max}}
\def\argmax{\textrm{argmax}}

\def\softmax{\textrm{softmax}}
\def\PLM{\textrm{PLM}}
\def\MLP{\textrm{MLP}}
\def\KL{\textrm{KL}}
\def\J{\textrm{J}}
\def\JS{\textrm{JS}}
\def\MSE{\textrm{MSE}}
\def\COS{\textrm{COS}}
\def\cos{\textrm{cos}}
\def\entropy{\textrm{H}}
\def\const{\textrm{constant}}
\def\ECE{\textrm{ECE}}
\def\SUP{\textsc{Sup}}
\def\REF{\textsc{Ref}}
\def\NEI{\textsc{Nei}}

\def\z{\textrm{z}}
\def\np{\textrm{np}}
\def\p{\textrm{p}}
\def\cp{\textrm{cp}}
\def\loss{L}
\def\reg{R}
\def\avgLoss{J}
\def\avgLossZ{\avgLoss_\z}
\def\avgLossNP{\avgLoss_\np}
\def\avgLossP{\avgLoss_\p}
\def\avgLossZeq{
\avgLossZ(\param) = \frac{1}{\dataSize}\sum_{(\inp,\lab)\in\dataSet}\loss(\inp,\lab;\param)
}
\def\lossCEeq{
\loss(\inp,\lab;\param) = \entropy(\trueDist,\predDist) = -\sum_{\lab\in\labSet}\trueDist(\lab|\inp)\log\predDist(\lab|\inp)
}
\def\avgLossNPeq{
\avgLossNP(\param) = \frac{1}{\dataSize_\np}\sum_{(\inp,\lab)\in\dataSet\cup\tranDataSet}\loss(\inp,\lab;\param)
}
\def\avgLossPeq{
\avgLossP(\param) = \frac{1}{\dataSize_\p}\;\sum_{t\in\langSet}\!\!\!\sum_{\substack{(\inp,\inpTran,\lab)\\\;\;\;\in(\dataSet,\tranDataSet_t)}}\!\!\!\loss(\inp,\inpTran,\lab;\param)
}
\def\lossPwithReq{
\loss(\inp,\inpTran,\lab;\param) = \loss(\inp,\lab;\param) + \loss(\inpTran,\lab;\param) + \lambda\reg(\param)
}
\def\regKLeq{
\reg(\param) = \KL(\predDist\parallel\predDistTran)
}
\def\regJeq{
\reg(\param) &= \J(\predDist\parallel\predDistTran)\nonumber\\
             &= \KL(\predDist\parallel\predDistTran) + \KL(\predDistTran\parallel\predDist)
}
\def\JSJbound{
\JS(\predDist\parallel\predDistTran) \leq \frac{1}{4}\J(\predDist\parallel\predDistTran)\nonumber
}
\def\meanDistTran{\frac{\predDist + \predDistTran}{2}}
\def\regJSeq{
\reg(\param) &= \JS(\predDist\parallel\predDistTran)\nonumber\\
             &= \frac{1}{2}\big(\KL(\predDist\parallel\meanDistTran) + \KL(\meanDistTran\parallel\predDistTran)\big)
}
\def\regMSEeq{
\reg(\param) = \|\inpVec - \inpTranVec\|^2
}
\def\regCOSeq{
\reg(\param) = 1 - \cos(\inpVec, \inpTranVec) = 1 - \frac{\inpVec\cdot\inpTranVec}{\|\inpVec\| \|\inpTranVec\|}
}
\def\lossCEasKLeq{
\loss(\inp,\lab;\param) &= \entropy(\trueDist,\predDist) - \entropy(\trueDist) +  \entropy(\trueDist)\nonumber\\
        &= \KL(\trueDist\parallel\predDist) + \underbrace{\entropy(\trueDist)}_{\const}
}
\def\lossCPeq{\loss(\inp,\lab;\param)_{\cp} = \entropy(\trueDist,\predDist) - \lambda\,\entropy(\predDist)}
\def\KLasHeq{%
\KL(\predDist\parallel\predDistTran) &= \entropy(\predDist,\predDistTran) - \entropy(\predDist),\nonumber\\
\KL(\predDistTran\parallel\predDist) &= \entropy(\predDistTran,\predDist) - \entropy(\predDistTran)
}
\def\lossJasCPeq{
\loss(\inp,\inpTran,\lab;\param) &= \entropy(\trueDist,\predDist) + \entropy(\trueDist,\predDistTran) + \lambda\J(\predDist\parallel\predDistTran)\nonumber\\
&= \entropy(\trueDist,\predDist) - \lambda\,\entropy(\predDist)\nonumber\\
&\;\;\;\;+ \entropy(\trueDist,\predDistTran) - \lambda\,\entropy(\predDistTran)\nonumber\\
&\;\;\;\;+ \lambda\big(\entropy(\predDist,\predDistTran) + \entropy(\predDistTran,\predDist)\big)
}

\title{XFEVER: Exploring Fact Verification across Languages}

\author{
Yi-Chen Chang$^1$\thanks{$\;\;$This work was conducted during the author's internship under National Institute of Informatics, Japan.} \hspace{2em} Canasai Kruengkrai$^2$ \hspace{2em} Junichi Yamagishi$^2$ \\
$^1$National Tsing Hua University, Taiwan \\
\texttt{yichen@nlplab.cc} \\
$^2$National Institute of Informatics, Japan \\
\texttt{\{canasai,jyamagishi\}@nii.ac.jp} }

\begin{document}

\maketitle
\begin{abstract}
This paper introduces the Cross-lingual Fact Extraction and VERification (XFEVER) dataset designed for benchmarking the fact verification models across different languages. 
We constructed it by translating the claim and evidence texts of the Fact Extraction and VERification (FEVER) dataset released by \citet{thorne-etal-2018-fever} into six languages. 
The training and development sets were translated using machine translation, whereas the test set includes texts translated by professional translators and machine-translated texts. 
Using the XFEVER dataset, two cross-lingual fact verification scenarios, \textit{zero-shot learning} and \textit{translate-train learning}, are defined, and baseline models for each scenario are also proposed in this paper.
Experimental results show that the multilingual language model can be used to build fact verification models in different languages efficiently. 
However, the performance varies by language and is somewhat inferior to the English case. 
We also found that we can effectively mitigate model miscalibration by considering the prediction similarity between the English and target languages.\footnote{The XFEVER dataset, code, and model checkpoints are available at \url{https://github.com/nii-yamagishilab/xfever}.}
\end{abstract}


\section{Introduction}

Automated fact verification is a part of the fact-checking task, verifying that a given claim is valid against a database of textual sources. 
It can be formulated as a classification task, taking the claim and associated evidence as input and determining whether the given evidence supports the claim. 
Deep learning is used to build classifiers for this purpose, but deep models are data-hungry and require massive amounts of labeled data. 
The Fact Extraction and VERification (FEVER) database~\cite{thorne-etal-2018-fever} is known as a well-resourced English database that enables us to build large networks, but building a database of the same scale as FEVER from scratch for each language is significantly time-consuming and costly. 
Our main question in this paper is: Can we build fact-checking models for other languages without huge costs?

\begin{figure*}
\begin{center}
\includegraphics[width=.68\textwidth]{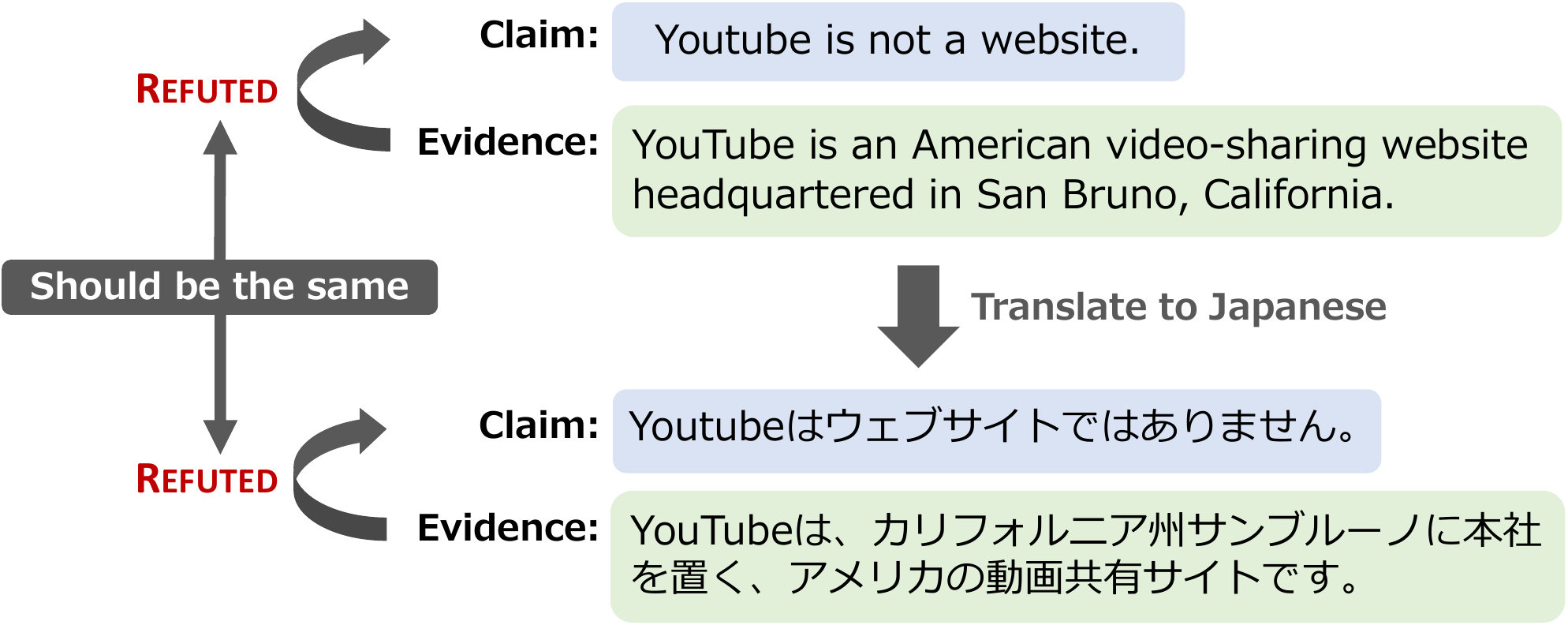}
\caption{
For the English example, it is clear that the given evidence refutes the claim.
Suppose we have {\em accurate} translations from English to another language (e.g., Japanese). 
The claim in Japanese must also be refuted on the basis of the evidence in Japanese.
In other words, the relationship between the claim and evidence text should be consistent across languages.
}\label{fig:ex-same-label}
\end{center}
\end{figure*}

In this work, we hypothesize that \textit{facts are facts regardless of languages}.
Suppose we have a perfect translator to translate English text into other languages without missing or changing information in the original texts. 
The relationship between a specific claim-evidence pair in the source language, which is the output of the fact verification model, should be the same even if they are translated into another target language as shown in Figure~\ref{fig:ex-same-label}. 
Using this hypothesis, we construct a new Cross-lingual Fact Extraction and VERification (XFEVER) dataset by automatically translating the claim and evidence texts of the FEVER dataset into five other languages: Spanish, French, Indonesian, Japanese, and Chinese. 
These languages cover several language families, including isolated languages such as Japanese.  
In addition to the machine-translated texts, a set of texts written and verified by professional translators is also available as an additional evaluation set to analyze whether the translation methods will affect the performance.

Using the XFEVER dataset, we define two cross-lingual fact verification scenarios: \textit{zero-shot learning} and \textit{translate-train learning}.
In the zero-shot learning scenario, the model is trained on the English corpus only and applied to other languages with zero shots.
In the translate-train learning scenario, a multilingual fact verification model is built in English and multiple languages, assuming that the machine-translated text in the non-English languages contains errors but is still somewhat useful for model training.
We also report baseline systems in each scenario.
In the zero-shot learning scenario, we show how beneficial the multilingual language models are. 
In the translate-train scenario, given the parallel data of texts translated from English into other languages, we also evaluate a baseline that uses the similarity of the predicted results or intermediate representations of the model in the English and other language cases as part of the loss.

The rest of the paper is organized as follows: We review the related work in the next section. 
Then, we overview the XFEVER dataset in Section~\ref{sec:xfever} and describe details of our baseline methods in Sections~\ref{sec:fact-verification} and~\ref{sec:consistency}. 
We provide experimental results in Section~\ref{sec:experiments}.
Finally, we summarize our research and future work in Section~\ref{sec:conclusion}.

\begin{table*}[t]
\small
\centering
{\tabcolsep = 0.4em
\begin{tabular}{cl}
\toprule
Language & Claim / Evidence\\
\midrule
\multirow{2}{*}{English} & Roman Atwood is a content creator. \B\\
                     & He is best known for his vlogs, where he posts updates about his life on a daily basis. \\
\midrule
\multirow{2}{*}{Spanish} & Roman Atwood es un creador de contenidos. \B\\
                       & Es conocido sobre todo por sus vlogs, en los que publica a diario noticias sobre su vida. \\
\midrule
\multirow{2}{*}{French} & Roman Atwood est un créateur de contenu. \B\\
                      & Il est surtout connu pour ses vlogs, où il publie quotidiennement des mises à jour sur sa vie. \\
\midrule
\multirow{2}{*}{Indonesian} & Roman Atwood adalah pembuat konten. \B\\
                         & Dia terkenal karena vlog-nya , di mana dia memposting pembaruan tentang hidupnya setiap hari.\\
\midrule
\multirow{2}{*}{Japanese} & {\scriptsize \begin{CJK}{UTF8}{ipxm}ローマン・アトウッドは、コンテンツクリエイター。\end{CJK}}\B\\
                        & {\scriptsize \begin{CJK}{UTF8}{ipxm}彼は彼のブログで最もよく知られている、彼は毎日のように彼の人生についての更新を投稿している。\end{CJK}}\\
\midrule
\multirow{2}{*}{Chinese} & {\scriptsize \begin{CJK}{UTF8}{gbsn}罗曼-阿特伍德是一个内容创作者。\end{CJK}}\B\\
                         & {\scriptsize \begin{CJK}{UTF8}{gbsn}他最出名的是他的博客，在那里他每天都会发布关于他的生活的更新。\end{CJK}}\\
\bottomrule
\end{tabular}
}
\caption{\label{tab:XFEVER-examples}
Examples (claim and evidence) from six languages in the XFEVER dataset with the $\SUP$ class.
}
\end{table*}

\section{Related Work}\label{sec:reelated-work}

\noindent\textbf{Automated fact-checking}

\noindent The importance of automated fact-checking is growing with an increase in misinformation, mal-information, and disinformation~\cite{Nakov21,Guo22}.
Automated fact-checking by machine learning, which should improve the efficiency of time-consuming fact-checking, consists of three steps~\cite{thorne-etal-2018-fever}: (1) searching the knowledge database to find out documents related to the claim to be verified, (2) finding sentences or paragraphs that serve as evidence in the documents found, and (3) predicting a verdict label for the claim to be verified on the basis of the retrieved evidence.

The third task, verdict prediction, is relevant to the textual entailment task~\cite{dagan_dolan_magnini_roth_2010} where using the given two sentences as inputs, we determine whether (i) they contradict each other or whether (ii) one sentence entails the other sentence without contradiction. 
The verdict prediction task examines whether the retrieved evidence entails the claim or whether they contradict each other.
Various architectures have been investigated, including graph-based neural networks~\cite{liu-etal-2020-fine, zhong-etal-2020-reasoning} and self-attention \cite{kruengkrai-etal-2021-multi}, and evaluations and comparisons have also been made using various language models \cite{lee-etal-2021-towards,Gopher21}.

\medskip\noindent\textbf{Fact-checking datasets}

\noindent There are several existing datasets for automated fact-checking.
FEVER~\cite{thorne-etal-2018-fever} and its series \cite{Thorne19FEVER2,Aly21Feverous} are well-known datasets for fact extraction and verification against textual sources. 
The original FEVER dataset consists of 185,445 claims manually verified against relevant Wikipedia articles.
WikiFactCheck~\cite{sathe-etal-2020-automated} is another dataset of 124K examples extracted from English Wikipedia articles and real-world claims (uncontrolled claims written by annotators).
Sources of evidence may change over time, requiring fact-checking models to be sensitive to subtle differences in supporting evidence. 
VitaminC~\cite{schuster-etal-2021-get} is a benchmark for testing whether a fact-checking model could identify such subtle factual changes.

\medskip\noindent\textbf{Datasets for cross-lingual understanding tasks}

\noindent Large multi-lingual language models such as mBERT~\cite{devlin2018bert} and XLM-R~\cite{conneau2019unsupervised} have been reported to be effective on cross-lingual tasks, and a number of benchmarks have been designed for the cross-lingual task: XTREME~\cite{pmlr-v119-hu20b} and XGLUE~\cite{liang-etal-2020-xglue}. 

The XTREME benchmark includes nine corpora and covers four natural language tasks: classification, structured prediction, question answering, and sentence retrieval. 
Among them, the Cross-lingual Natural Language Inference (XNLI) corpus~\cite{conneau-etal-2018-xnli} is the most related to XFEVER, which is an extended version of the Multi-Genre Natural Language Inference (MultiNLI) corpus~\cite{williams-etal-2018-broad} and contains 15 languages translated by professional translators.
There exists a multilingual fact-checking dataset named X-FACT, which consists of 31,189 real-world claims collected from fact-checking websites~\cite{Gupta21}.
Although XNLI (and our XFEVER) can be regarded as artificially created datasets, they have certain advantages, such as having similar data distributions across languages~\cite{conneau-etal-2018-xnli}.

\begin{table}[t]
    \small
    \begin{center}
    {
    \begin{tabular}{llrrr}
    \toprule
    Split & Trans & {\centering$\SUP$} & $\REF$ & $\NEI$ \\
    \midrule
    Train & Machine & 100,570 & 41,850 & 35,639\B\\
    Dev   & Machine & 3,964 & 4,323 & 3,333\B\\
    Test  & Machine & 4,019 & 4,358 & 3,333\B\\
    \multirow{2}{*}{Test-6h} & Machine & 200 & 200 & 200\\
          & Human & 200 & 200 & 200\\
    \bottomrule
    \end{tabular}
    }
    \caption{
    Number of examples per class for each target language in the XFEVER dataset. 
    The column ``Trans'' indicates the translation method.
    The test-6h set consists of two small subsets: machine- and human-translated sets.
    }
    \label{XFEVER-statistics}
    \end{center}
\end{table}

\section{The XFEVER dataset}\label{sec:xfever}

\subsection{Overview}

Inspired by the XNLI dataset construction~\cite{conneau-etal-2018-xnli}, we extended the FEVER dataset~\cite{thorne-etal-2018-fever} to XFEVER by translating the English claim-evidence pairs into different languages. 
We used the dataset version pre-processed by~\citet{schuster-etal-2021-get}, where only claims that require evidence from single sentences are considered.
We considered a total of six languages: Spanish (es), French (fr), Indonesian (id), Japanese (ja), Chinese (zh), and the source language English (en).

Table~\ref{tab:XFEVER-examples} shows examples in the languages included in the XFEVER dataset.
We automatically translated the original English data to the five target languages using DeepL.\footnote{\url{https://www.deepl.com/pro-api}}
To analyze whether the translation methods affect the prediction accuracy, we created a small test set (test-6h) containing 600 randomly-selected claim-evidence pairs translated and verified by professional translators.

Table~\ref{XFEVER-statistics} shows the data statistics per language.
Each claim-evidence pair has one of the class labels: supported ($\SUP$), refuted ($\REF$), and not enough info ($\NEI$). 
We assigned the same labels as the original ones to translated pairs.

\subsection{Two scenarios}
Given the XFEVER dataset, we explore two scenarios.

\begin{itemize}[leftmargin=1.2em]
\item \textbf{Zero-shot learning:}
We can only access the English training and development sets to train a model and evaluate the trained model on the test set in all languages.
\item \textbf{Translate-train learning:}
We assume that machine-translated data are available.
We then build a model using the training and development sets in all languages simultaneously. 
The evaluation is the same as the zero-shot learning scenario.
\end{itemize}

\section{Cross-lingual fact verification} \label{sec:fact-verification}

In this section, we first introduce notation and then describe the frameworks for zero-shot and translate-train learning scenarios.
We consider cross-lingual fact verification as a classification problem.
We want to train a model $\modelP: \inpSet \rightarrow \labSet$ parameterized by $\param$, which maps an input $\inp \in \inpSet$ to a label $\lab \in \labSet = \{1,\ldots,\nCls\}$.\footnote{In our task, $\nCls = 3$, where $1 = \SUP$, $2 = \REF$, and $3 = \NEI$.}
Our model is a neural network consisting of a multilayer perceptron ($\MLP$) on top of a pre-trained language model ($\PLM$):
\begin{align}
\modelPeq\nonumber.
\end{align}
The $\PLM$ takes $\inp$ (a concatenation of claim and evidence sentences) as input and produces a vector representation.
The $\MLP$ then maps the vector representation to $\nCls$ real-valued numbers (i.e., logits).
We finally obtain the predicted probability $\predDist \in \Real^\nCls$ by applying the $\softmax$ function:
\begin{align}\label{eq:pred-dist}
\predDistEq.
\end{align}

\subsection{Zero-shot learning scenario}

In the \textit{zero-shot learning} scenario, we only use the original data $\trainSet$ for training.
In our study, we refer to the original data as the non-translated data, which are in English.
We aim to minimize the average loss:
\begin{align}\label{eq:avg-loss-zero}
\avgLossZeq,
\end{align}
where the loss function $\loss(\inp,\lab;\param)$ is the cross-entropy between the ground-truth label distribution $\trueDist \in \Real^\nCls$ (i.e., one-hot encoding) and the predicted distribution $\predDist$:
\begin{align}\label{eq:loss-ce}
\lossCEeq.
\end{align}
With help from the multilingual $\PLM$ (e.g., mBERT or XML-R), we expect that the zero-shot model would work with other languages as well.

\subsection{Translate-train learning scenario}

In the \textit{translate-train learning} scenario, we assume that the machine-translated data $\tranDataSet$ exists so that we can exploit them for training.
We define $\tranDataSetEq$, where $\langSetEq$ is the set of our target languages.

\subsubsection{Non-parallel training}

The most straightforward strategy is to mix all the available data.
We write the average loss for non-parallel ($\np$) training as:
\begin{align}\label{eq:avg-loss-np}
\avgLossNPeq,
\end{align}
where $\medmuskip=0mu\dataSize_\np = \dataSize\times(|\langSet|+1)$ is the number of all mixed examples.
The loss function $\loss(\inp,\lab;\param)$ is still the cross-entropy loss.
In practice, we reshuffle the training examples at the beginning of each epoch, so $\inp$ comes from $\dataSet$ or $\tranDataSet$ at random.

\subsubsection{Parallel training}

Non-parallel training does not consider that the predicted label of the machine-translated example $\inpTran$ should be the same as the original example $\inp$.
To take the consistency of predictions into account, we explicitly create parallel examples of $\inp$ and $\inpTran$ and use such pairs for training.
We formulate the average loss for parallel ($\p$) training as:

\begin{align}\label{eq:avg-loss-p}
\avgLossPeq,
\end{align}
where $\medmuskip=0mu\dataSize_\p = \dataSize\times|\langSet|$ is the number of all parallel examples.
Since we reshuffle parallel examples at every epoch similar to non-parallel training, $\inpTran$ comes from one of $\tranDataSet_t$ randomly.
We define the loss function $\loss(\inp,\inpTran,\lab;\param)$ as:
\begin{align}\label{eq:loss-p-r}
\lossPwithReq,
\end{align}
where the first and second terms are the cross-entropy losses for the original and translated examples, and the last term $\reg(\param)$ is a regularization function with a strength coefficient $\lambda$.
In the following section, we discuss various choices for $\reg(\param)$.

\section{Consistency regularization}\label{sec:consistency}

We use the regularization function $\reg(\param)$ to enforce cross-lingual consistency.
Previous work has presented specific forms of consistency regularization~\cite{Zheng21,Yang22}.
Here, we examine a wide range of regularization functions where we categorize them into types: prediction and representation.
In addition, we discuss how prediction consistency relates to the confidence penalty.

\subsection{Prediction consistency}\label{sec:pred-consistency}

Let $\predDistTran(\lab|\inpTran)$ denote the predicted distribution given the machine-translated example $\inpTran$.
Intuitively, the predicted distributions for the original and translated examples should be close to reaching the same predictions.
To achieve this, we can regularize the loss in Eq.~(\ref{eq:loss-p-r}) with an information-theoric divergence measure between $\predDist$ and $\predDistTran$.
We explore the following divergence measures:

\begin{itemize}[leftmargin=1.2em]
\item \textbf{Kullback–Leibler ($\KL$) divergence}:
We hypothesize that the prediction of the original example tends to have better accuracy than the machine-translated one.
Thus, we push $\predDistTran$ towards $\predDist$ with the $\KL$ divergence~\cite{Kullback51}:
\begin{align}\label{eq:reg-kl}
\regKLeq.
\end{align}
\item \textbf{Jeffreys ($\J$) divergence}:
The multilingual information in the $\PLM$ can be helpful and captured through the translated example.
Also, to promote the consistency of predictions, we push $\predDist$ and $\predDistTran$ towards each other by applying the symmetric measure called the J divergence~\cite{Jeffreys46}:
\begin{align}\label{eq:reg-j}
\regJeq.
\end{align}
\item \textbf{Jensen–Shannon ($\JS$) divergence}:
The $\KL$ and $\J$ divergence measures are unbound.
Another symmetric and bounded measure is the $\JS$ divergence~\cite{Lin91}:
\begin{align}\label{eq:reg-js}
\regJSeq.
\end{align}
\end{itemize}

\noindent\textbf{Relationship between prediction consistency and confidence penalty}

\noindent When the model predicts a label with a probability (i.e., confidence) of 0.95, we expect it to have a 95\% chance of being correct.
However, researchers have found that neural models tend to be overconfident.
In other words, the model's confidence poorly aligns with the ground-truth correctness likelihood.
\citet{Guo17} attributed the cause of overconfident predictions to cross-entropy loss overfitting, where the model places most of the probability mass on a single label, resulting in a peaked predicted distribution.

In this section, we discuss cross-entropy loss overfitting from a KL divergence perspective.
We can rewrite the cross-entropy loss in Eq.~(\ref{eq:loss-ce}) in a KL divergence form as:
\begin{align}
\lossCEasKLeq\nonumber.
\end{align}
Thus, we minimize the loss at training time by pushing $\predDist$ (the predicted distribution) towards $\trueDist$ (the ground-truth one-hot distribution). 
When overfitting occurs, $\predDist$ becomes peaky.

There are several calibration methods to mitigate the above issue.
One of which is the confidence penalty~\cite{Pereyra17} in which a penalized term (i.e., a negative entropy) is added to the cross-entropy loss:
\begin{align}
\lossCPeq\nonumber.
\end{align}
The model attempts to maximize the entropy $\entropy(\predDist)$ to minimize the loss $\loss(\inp,\lab;\param)_\cp$.
Thus, $\predDist$ becomes smoother (or less peaky).

Our key observation is that the regularization functions of prediction consistency intrinsically introduce the confidence penalty to the loss.
Let us consider the parallel training loss with the $\J$ divergence as an example.
We know that:
\begin{align}
\KLasHeq\nonumber.
\end{align}
From Eqs.~(\ref{eq:loss-ce}),~(\ref{eq:loss-p-r}), and~(\ref{eq:reg-j}), we obtain:
\begin{align}\label{eq:loss-j-as-cp}
\lossJasCPeq.
\end{align}
Thus, the loss in Eq.~(\ref{eq:loss-j-as-cp}) includes the negative entropy terms of $\predDist$ and $\predDistTran$, which should help reduce model overconfidence.
We verify this observation in Section~\ref{sec:res-ece}.

\subsection{Representation consistency}\label{sec:repr-consistency}

Recall that we derive the predicted distribution from the logits in Eq.~(\ref{eq:pred-dist}).
We can also impose consistency in the intermediate representation before the logits. 
Here, we examine two representation levels: penultimate and feature.
We refer to the penultimate and feature representations as the output of the last layer right before the logits and that of the PLM, respectively.
Let $\inpVec$ and $\inpTranVec$ be the representations\footnote{They can be either penultimate or feature representation.} of the original and translated examples.
Since both representations are vectors, we can apply the following distance measure:

\begin{itemize}[leftmargin=1.2em]
\item \textbf{Mean square error (MSE)}: We compute the MSE (or the square of Euclidean distance) as:
\begin{align}\label{eq:reg-mse}
\regMSEeq.
\end{align}
Thus, if $\inpVec$ and $\inpTranVec$ are similar, $\reg(\param)$ approaches zero.
\item \textbf{Cosine distance (COS)}: An alternative measure is the cosine distance computed as:
\begin{align}\label{eq:reg-cos}
\regCOSeq.
\end{align}
For the cosine distance, the magnitudes of $\inpVec$ and $\inpTranVec$ have no effect because they are normalized to the unit vectors.
\end{itemize}

\section{Experiments} \label{sec:experiments}

\subsection{Training details}

We implemented our models using Hugging Face's Transformers library~\cite{Wolf20}.
In the zero-shot setting, we compared the multilingual $\PLM$s against their monolingual versions to examine their benefits.
For the monolingual $\PLM$s, we used BERT-base (110M), RoBERTa-base (125M), and RoBERTa-large (355M).
The number in the parenthesis denotes the number of parameters.
For the multilingual $\PLM$s, we used mBERT (178M), XLM-R-base (470M), and XLM-R-large (816M).
The mBERT model was pre-trained on the Wikipedia entries of 104 languages, while the XLM-R models were pre-trained on the Common Crawl Corpus covering 100 languages. 
The pre-training datasets for mBERT and XLM-R include all six languages in the XFEVER dataset.

For all experiments, we used the Adafactor optimizer~\cite{Shazeer18ada} with a batch size of 32. 
We used a learning rate of 2e-5 for BERT-base/RoBERTa-base/mBERT and 5e-6 for RoBERTa-large/XLM-R-large.
We trained each model for up to ten epochs or until the accuracy on the development set had not improved for two epochs.
For consistency regularization, we set $\lambda$ to 1 unless otherwise specified.
We conducted all the experiments on 32GB NVIDIA Tesla A100 GPUs.

\subsection{Results}\label{sec:results}

\begin{table*}[t]
\small
\begin{center}
\begin{tabular}{lccccccc}
\toprule
$\PLM$ & en & es & fr & id & ja & zh & Avg \\
\midrule
{\em Monolingual} \\
BERT & 87.7 & 53.2 & 53.2 & 49.6 & 36.9 & 39.1 & 53.3 \\
RoBERTa-base & 88.9 & 67.4 & 67.2 & 56.5 & 40.3 & 37.7 & 59.7 \\
RoBERTa-large & \bf90.1 & 79.2 & 72.2 & 54.3 & 39.0 & 37.5 & 62.1 \\
\midrule
{\em Multilingual} \\
mBERT & 87.9 & 83.7 & 84.3 & 82.6 & 72.4 & 82.1 & 82.2 \\
XLM-R-base & 87.7 & 83.7 & 81.3 & 81.9 & 74.4 & 78.0 & 81.2 \\
XLM-R-large & 89.5 & \bf87.3 & \bf85.3 & \bf85.5 & \bf82.0 & \bf83.1 & \bf85.5 \\
\bottomrule
\end{tabular}
\caption{
Accuracy scores of monolingual and multilingual $\PLM$s on the test set in zero-shot learning $\avgLossZ$.
}\label{tab:results-zero-shot}
\end{center}
\end{table*}

\begin{table*}[t]
\small
\begin{center}
\begin{tabular}{lllccccccc}
\toprule
Model & Consistency & $\reg$ & en & es & fr & id & ja & zh & Avg \\
\midrule
Zero-shot $\avgLossZ$     & -- & -- & 87.9 & 83.7 & 84.3 & 82.6 & 72.4 & 82.1 & 82.2\B\\
Non-parallel $\avgLossNP$ & -- & -- & \bf88.1 & \bf86.8 & \bf86.5 & 86.0 & \bf85.4 & \bf86.0 & \bf86.5\B\\
Parallel $\avgLossP$      & -- & -- & 87.0 & 85.7 & 85.7 & 85.3 & 79.8 & 82.9 & 84.4\B\\    
                     & Pred & $\KL$ & 87.4 & 86.1 & 85.7 & 85.6 & 81.4 & 84.1 & 85.0 \\
                     &       & $\J$ & 86.9 & 85.7 & 85.6 & 85.8 & 81.7 & 83.9 & 84.9 \\
                     &       & $\JS$ & 87.4 & 86.0 & 85.8 & 85.9 & 81.7 & 84.2 & 85.2\B\\
                     & Repr & $\MSE$-feat & 87.4 & 85.7 & 86.0 & 85.9 & 82.2 & 85.1 & 85.4 \\
                     &      & $\MSE$-penu & 87.5 & 86.1 & 86.0 & \bf86.2 & 82.4 & 84.4 & 85.4 \\
                     &      & $\COS$-feat & 87.4 & 85.7 & 85.8 & 85.8 & 83.0 & 84.3 & 85.3 \\
                     &      & $\COS$-penu & 87.1 & 85.7 & 85.7 & 85.7 & 82.2 & 84.1 & 85.1 \\
\bottomrule
\end{tabular}
\caption{
Accuracy scores of mBERT on the test set.
Pred = Prediction; Repr = Representation;
feat = feature; penu = penultimate.
}\label{tab:results-translate-train}
\end{center}
\end{table*}

\begin{table}[t]
\small
\begin{center}
\begin{tabular}{lcc}
\toprule
Consistency ($\reg$) & mBERT & XLM-R-large \\
\midrule
-- & 84.4 & \bf88.3 \\
Pred ($\JS$) & 85.2 & 88.1 \\
Repr ($\MSE$-feat) & \bf85.4 & 88.1 \\
Pred ($\JS$) \& Pepr ($\MSE$-feat) & 85.3 & 88.0 \\
\bottomrule
\end{tabular}
\caption{
Additional results of parallel training $\avgLossP$.
}\label{tab:results-combine}
\end{center}
\end{table}

\begin{table*}[t]
\small
\begin{center}
\begin{tabular}{llllccccccc}
\toprule
Model & Consistency & $\reg$ & en & es & fr & id & ja & zh & Avg \\
\midrule
Zero-shot $\avgLossZ$     & -- & -- & 6.0 & 8.5 & 7.9 & 9.2 & 14.6 & 8.6 & 9.1\B\\
Non-parallel $\avgLossNP$ & -- & -- & 4.9 & 5.2 & 5.2 & 5.4 & 4.2 & 5.0 & 5.0\B\\
Parallel $\avgLossP$      & -- & -- & 8.7 & 7.5 & 7.4 & 7.7 & 7.6 & 6.2 & 7.5\B\\
                     & Pred & $\KL$ & 3.4 & 5.2 & 5.6 & 5.8 & 8.4 & 6.4 & 5.8 \\
                     &      & $\J$ & \bf1.5 & \bf2.4 & \bf2.7 & \bf2.6 & 5.3 & 4.1 & \bf3.1 \\
                     &      & $\JS$ & 3.5 & 3.1 & \bf2.7 & 2.8 & \bf4.1 & \bf3.8 & 3.3\B\\
                     & Repr & $\MSE$-feat & 8.1 & 8.3 & 7.9 & 8.0 & 7.6 & 6.7 & 7.8 \\
                     &      & $\MSE$-penu & 7.6 & 7.2 & 7.2 & 7.2 & 6.5 & 6.3 & 7.0 \\
                     &      & $\COS$-feat & 8.7 & 8.6 & 8.5 & 8.2 & 7.7 & 7.3 & 8.2 \\
                     &      & $\COS$-penu & 8.9 & 8.1 & 8.0 & 8.2 & 8.0 & 7.8 & 8.2 \\
\bottomrule
\end{tabular}
\caption{
$\ECE$ scores (lower is better) of mBERT on the test set.
}
\label{tab:results-ece}
\end{center}
\end{table*}

\begin{table*}[t]
\small
\begin{center}
\begin{tabular}{lllcccccc}
\toprule
Scenario & $\PLM$ & Trans & es & fr & id & ja & zh & Avg \\
\midrule
Zero-shot $\avgLossZ$ & mBERT & Machine & 83.5 & 83.8 & 82.3 & 74.3 & 82.5 & 81.3 \\
                &             & Human   & 83.5 & 84.8 & 81.5 & 77.2 & 83.0 & 82.0\B\\
                & XLM-R-large & Machine & 85.2 & 83.3 & 85.0 & 81.3 & 83.5 & 83.7 \\
                &             & Human   & 83.8 & 84.2 & 83.3 & 83.7 & 82.0 & 83.4 \\
\midrule
Translate-train $\avgLossNP$ & mBERT & Machine & 87.2 & 85.8 & 87.2 & 83.5 & 85.8 & 85.9 \\
                       &             & Human   & 87.5 & 86.7 & 86.2 & 82.0 & 84.8 & 85.4\B\\
                       & XLM-R-large & Machine & 86.8 & 86.7 & 87.5 & 86.2 & 87.2 & 86.9 \\
                       &             & Human   & 86.0 & 87.0 & 85.5 & 87.7 & 84.7 & 86.2 \\
\bottomrule
\end{tabular}
\caption{
Comparison of accuracy scores on the machine- and human-translated test-6h set.
}
\label{tab:results-machine-human}
\end{center}
\end{table*}

\subsubsection{Effect of multilingual $\PLM$s in zero-shot learning}\label{sec:res-zero-shot}

Table~\ref{tab:results-zero-shot} shows the accuracy gains of multilingual $\PLM$s over the monolingual counterparts in the zero-shot learning scenario.
Specifically, we obtain +28.9\% (BERT$\rightarrow$mBERT), +21.5\% (RoBERTa-base$\rightarrow$XLM-R-base), and +23.4\% (RoBERTa-large$\rightarrow$XLM-R-large) improvements on average.
As expected, the monolingual $\PLM$s yield high accuracy for the source language (English) but cannot maintain reasonable accuracy for the target languages.
The multilingual $\PLM$s help alleviate this issue.
For example,  changing RoBERTa-large$\rightarrow$XLM-R-large yields +43\% and +45.6\% improvements for Japanese and Chinese, respectively.
These results indicate that the multilingual $\PLM$s are extremely helpful when the training set in the target language are unavailable.

\subsubsection{Effect of translate-train learning on performance improvement}\label{sec:res-translate-train}

Table~\ref{tab:results-translate-train} shows the results of various settings using mBERT.\footnote{The results of XLM-R-large are in Appendix~\ref{appendix:additional-results}.}
When we can access machine-translated data, our non-parallel training $\avgLossNP$ works well for most target languages.
The type of regularization functions or representations has less effect on performance in terms of accuracy.
As shown in Table~\ref{tab:results-combine}, we also attempt to combine prediction and representation consistencies.
While these consistencies improve the accuracy scores with mBERT, their effects diminish with XLM-R-large. 
In the next section, we inspect the benefit of consistency regularization in reducing miscalibration.

\subsubsection{Effect of consistency regularization in reducing miscalibration}\label{sec:res-ece}

We can quantify miscalibration by measuring the gap between model confidence (conf) and accuracy (acc).
A common metric is the expected calibration error ($\ECE$,~\citealt{Naeini15}):
\begin{align}
\ECEeqs
\end{align}
where $\binSet_\binId$ is the set of examples belonging to the $\binTh$ bin.\footnote{We divide the confidence range of $[0, 1]$ into $\binSize$ equal-size bins, where the $\binTh$ bin covers the interval of $\binInterval$. We set $\binSize=20$.
}

In Section~\ref{sec:pred-consistency}, we find that our prediction consistency contains the negative entropy of the predicted distribution, which should help mitigate miscalibration as in the confident penalty~\cite{Pereyra17}.
As shown in Table~\ref{tab:results-ece}, the symmetric divergence measures, $\J$ and $\JS$, significantly reduce the ECE scores because they encourage the model to output high entropy for both the original and translated examples.
Although we observed slight differences in accuracy among our regularization functions in Section~\ref{sec:res-translate-train}, we would prefer a model having lower $\ECE$ (i.e., better calibrated) in practice.
Thus, we suggest applying prediction consistency with a symmetric divergence measure ($\J$ or $\JS$).

\subsubsection{Performance comparison of human- and machine-translated data}\label{sec:machine-human}

So far, we have used machine-translated data to evaluate the performance on the target languages.
We now examine whether there is a performance disparity between machine- and human-translated data because we expect to apply our model to human-written texts.
We experiment with the test-6h set, where a subset of 600 examples from the original test set were translated by both machines (DeepL) and professional translators.

As shown in Table~\ref{tab:results-machine-human}, the average differences are only around 0.3$\sim$0.7\%.
We attribute these minor discrepancies to DeepL's accurate translations.
Our results suggest that translate-train learning is effective when we can have high-quality translated data. 
Appendix~\ref{appendix:auto-human-examples} shows examples of the machine- and human-translated texts from the test-6h set.

\section{Conclusion}\label{sec:conclusion}

False claims can spread across languages.
Identifying these claims is an important task since a number of online claims might cause harm in the real world.
Existing benchmarks for fact verification are mainly in English.
To address the lack of benchmarks for non-English languages, we introduced the XFEVER dataset for the cross-lingual fact verification task.

We presented a series of baselines in two scenarios: zero-shot learning and translate-train learning. 
For the latter scenario, we explored various regularization functions.
We found that translate-train learning with high-quality machine-translated data can be effective.
In addition, consistency regularization with symmetric divergence measures can help reduce miscalibration.

For future work, we plan to investigate a scenario when large machine-translated data are unavailable, but we can acquire a few examples for training. 
We also want to expand XFEVER's human-translated data to cover more languages, especially low-resource ones.

\section*{Acknowledgments}
This work is supported by JST CREST Grants (JPMJCR18A6 and JPMJCR20D3) and MEXT KAKENHI Grants (21H04906), Japan.

\bibliography{main}
\bibliographystyle{acl_natbib}

\appendix

\section{Additional results}\label{appendix:additional-results}

We conducted preliminary experiments and found that the default $\lambda = 1$ does work well with the $\J$ divergence and XLM-R-large.
One plausible reason is that the $\J$ divergence penalizes the loss more heavily than other divergence measures.
If we follow the proof of Theorem 1 in~\citet{Lin91}, we can obtain the following bound:
\begin{align}\label{eq:reg-mse}
\JSJbound.
\end{align}
Thus, we heuristically reduce $\lambda$ to $0.25$ for the $\J$ divergence to alleviate the issue.
Tables~\ref{tab:results-translate-train-xlm} and~\ref{tab:results-ece-xlm} show the accuracy and $\ECE$ scores of XLM-R-large on the test set, respectively.

\section{Machine vs. human translations}\label{appendix:auto-human-examples}

Table~\ref{tab:machine-human} shows examples of the machine- and human-translated texts from the test-6h set.  

\begin{table*}[t]
\small
\begin{center}
\begin{tabular}{lllccccccc}
\toprule
Model & Consistency & $\reg$ & en & es & fr & id & ja & zh & Avg \\
\midrule
Zero-shot $\avgLossZ$     & -- & -- & 89.5 & 87.3 & 85.3 & 85.5 & 82.0 & 83.1 & 85.5\B\\
Non-parallel $\avgLossNP$ & -- & -- & \bf89.7 & \bf88.7 & \bf88.4 & 88.4 & \bf88.1 & \bf88.0 & \bf88.6\B\\
Parallel $\avgLossP$      & -- & -- & 89.7 & 88.5 & 87.6 & 88.7 & 87.4 & 87.7 & 88.3\B\\
                     & Pred & $\KL$ & 89.3 & 88.4 & 87.1 & 88.4 & 86.8 & 87.1 & 87.8 \\
                     &      & $\J$ & 89.6 & 88.5 & 87.7 & \bf88.8 & 87.1 & 87.7 & 88.2 \\
                     &      & $\JS$ & \bf89.7 & 88.3 & 87.4 & 88.4 & 87.1 & 87.6 & 88.1\B\\
                     & Repr & $\MSE$-feat & \bf89.7 & 88.4 & 87.5 & 88.7 & 87.0 & 87.5 & 88.1 \\
                     &      & $\MSE$-penu & \bf89.7 & 88.5 & 87.6 & 88.4 & 86.7 & 87.7 & 88.1 \\
                     &      & $\COS$-feat & 89.5 & 88.4 & 87.6 & 88.5 & 87.4 & 87.5 & 88.1 \\
                     &      & $\COS$-penu & 89.6 & 88.4 & 87.5 & 88.4 & 87.0 & 87.6 & 88.1 \\
\bottomrule
\end{tabular}
\caption{
Accuracy scores of XLM-R-large on the test set.
Pred = Prediction; Repr = Representation;
feat = feature; penu = penultimate.
}\label{tab:results-translate-train-xlm}
\end{center}
\end{table*}

\begin{table*}[t]
\small
\begin{center}
\begin{tabular}{llllccccccc}
\toprule
Model & Consistency & $\reg$ & en & es & fr & id & ja & zh & Avg \\
\midrule
Zero-shot $\avgLossZ$     & -- & -- & 8.8 & 10.6 & 12.4 & 12.4 & 15.1 & 14.2 & 12.2\B\\
Non-parallel $\avgLossNP$ & -- & -- & 6.0 & 6.5 & 6.6 & 6.9 & 5.9 & 6.5 & 6.4\B\\
Parallel $\avgLossP$      & -- & -- & 5.7 & 5.3 & 5.3 & 5.4 & 3.7 & 4.6 & 5.0\B\\
                     & Pred & $\KL$ & \bf2.4 & 4.0 & 5.0 & 4.3 & 4.9 & 5.0 & 4.3 \\
                     &      & $\J$ & 3.6 & 4.4 & 4.5 & 4.4 & 4.2 & 4.5 & 4.3 \\
                     &      & $\JS$ & 2.6 & \bf2.8 & \bf2.9 & \bf2.8 & \bf3.1 & \bf2.7 & \bf2.8\B\\
                     & Repr & $\MSE$-feat & 4.8 & 4.8 & 5.0 & 4.9 & 3.8 & 4.5 & 4.6 \\
                     &      & $\MSE$-penu & 5.5 & 5.6 & 5.9 & 6.1 & 5.3 & 5.6 & 5.7 \\
                     &      & $\COS$-feat & 5.3 & 5.4 & 5.5 & 5.7 & 4.4 & 5.3 & 5.3 \\
                     &      & $\COS$-penu & 5.8 & 5.7 & 5.8 & 5.9 & 4.7 & 5.3 & 5.5 \\
\bottomrule
\end{tabular}
\caption{
$\ECE$ scores (lower is better) of XLM-R-large on the test set.
}
\label{tab:results-ece-xlm}
\end{center}
\end{table*}

\begin{table*}[t]
\small
\begin{center}
\begin{tabular}{cll}
\toprule
Language & Trans & Claim / Evidence\\
\midrule
\multirow{2}{*}{English} & \multirow{2}{*}{Original} & Simon Pegg is an actor. \\
                         & & He and Nick Frost wrote and starred in the sci-fi film Paul ( 2011 ). \\
\midrule
\multirow{4}{*}{Spanish}  & \multirow{2}{*}{Machine} & Simon Pegg es un actor. \\
                         & & Él y Nick Frost escribió y protagonizó la película de ciencia ficción Paul ( 2011 ).\B\\
                          & \multirow{2}{*}{Human} & Simon Pegg es un actor.\\
                         & & Él y Nick Frost escribieron y protagonizaron la película de ciencia ficción Paul (2011). \\
\midrule
\multirow{2}{*}{French} & \multirow{2}{*}{Machine} & Simon Pegg est un acteur. \\
                      & & Avec Nick Frost, il a écrit et joué dans le film de science-fiction Paul ( 2011 ).\B\\
                        & \multirow{2}{*}{Human} & Simon Pegg est un acteur.\\
                      & & Avec Nick Frost, il a écrit et joué dans le film de science-fiction Paul (2011). \\
\midrule
\multirow{2}{*}{Japanese} & \multirow{2}{*}{Machine} & \begin{CJK}{UTF8}{ipxm}サイモン・ペッグは、俳優である。\end{CJK} \\
                        & & \begin{CJK}{UTF8}{ipxm}ニック・フロストとともにSF映画『ポール』( 2011 ) で脚本と主演を務めた。\end{CJK}\B\\
                        & \multirow{2}{*}{Human} & \begin{CJK}{UTF8}{ipxm}Simon Peggは俳優です。\end{CJK}\\
                      & & \begin{CJK}{UTF8}{ipxm}彼と Nick FrostはSF映画『Paul』(2011年)の脚本を書き、主演もしています。\end{CJK} \\
\midrule
\multirow{2}{*}{Chinese} & \multirow{2}{*}{Machine} & \begin{CJK}{UTF8}{gbsn}西蒙-佩吉是一名演员。\end{CJK} \\
                         & & \begin{CJK}{UTF8}{gbsn}他和尼克-弗罗斯特编剧并主演了科幻电影《保罗》(2011)。\end{CJK}\B\\
                         & \multirow{2}{*}{Human} & \begin{CJK}{UTF8}{gbsn}西蒙·佩吉是一名演员。\end{CJK}\\
                         & & \begin{CJK}{UTF8}{gbsn}他和尼克·弗罗斯特(Nick Frost)在科幻电影《保罗》(2011)中担任编剧并主演。\end{CJK} \\
\bottomrule
\end{tabular}
\caption{
Examples (claim and evidence) from six languages in the XFEVER's test-6h set. Machine = DeepL; Human = professional translators.
}\label{tab:machine-human}
\end{center}
\end{table*}

\end{document}